\documentclass[runningheads]{llncs}

\usepackage{eccv}

\usepackage{eccvabbrv}

\usepackage{graphicx}
\usepackage{booktabs}
\usepackage{multirow}
\usepackage{colortbl} 
\usepackage{tcolorbox}
\usepackage[misc]{ifsym}
\usepackage[accsupp]{axessibility}  

\definecolor{mygray}{gray}{.92}
\makeatletter
\def\blfootnote{\xdef\@thefnmark{}\@footnotetext}
\makeatother

\usepackage{hyperref}

\usepackage{orcidlink}

\begin{document}

\title{AdaThinking-\includegraphics[height=1.5em]{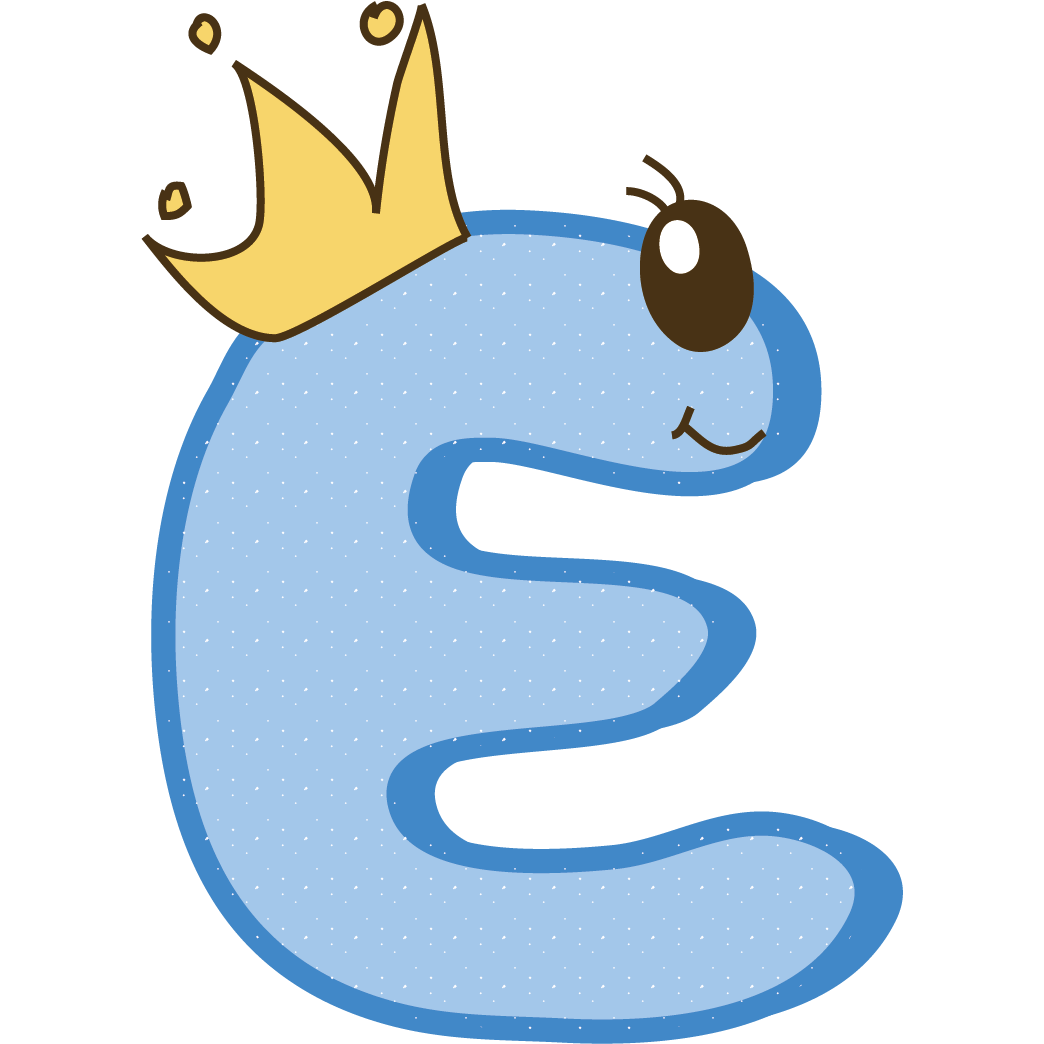}: One-Token Entropy Regulation for Adaptive Thinking} 

\titlerunning{AdaThinking-\includegraphics[height=1.5em]{fig/E.png}}

\author{Zining Wang\inst{1}\textsuperscript{$*$}\orcidlink{0009-0007-3847-6396} \and
Tongkun Guan\inst{2}\textsuperscript{$*$}\orcidlink{0000-0003-3346-8315} \and
Boming Chen\inst{1}\textsuperscript{$*$}\orcidlink{0009-0003-1248-2430} \and
Zhentao Guo\inst{1}\orcidlink{0009-0000-0242-4059} \and Jianqiang Liu\inst{1} \and Chao Jin\inst{3} \and Chen Duan\inst{1} \and Kai Zhou\inst{1(\textrm{\Letter})} \and Pengfei Yan\inst{1} \and Wei Shen\inst{2(\textrm{\Letter})} \and  Xiaokang Yang\inst{2}}

\authorrunning{Z.~Wang et al.}

\institute{Meituan, \email{\{wangzining03,chenboming,guozhentao,zhoukai03\}@meituan.com}
\and
MoE Key Lab of Artificial Intelligence, AI Institute, School of Computer Science, \\ Shanghai Jiao Tong University, \email{gtk0615@sjtu.edu.cn} \and
MAIS\&NLPR, Institute of Automation, Chinese Academy of Sciences
}

\maketitle
\blfootnote{\noindent$^{*}$Equal contribution. \textsuperscript{\Letter}Corresponding author.}
\begin{abstract}
Multimodal large language models have demonstrated strong document reasoning capabilities by incorporating explicit thinking processes. While this capability significantly improves performance on challenging tasks, current models apply such deep reasoning uniformly to all questions, resulting in unnecessary computational overhead for simple task. This not only degrades user experience but also negatively impact accuracy on benchmark datasets. We identify the critical need for adaptive thinking mechanisms that can intelligently determine when to engage reasoning based on question complexity. To address this, we propose AdaThinking-E, a novel reinforcement learning framework that learns adaptive thinking through one-token entropy regulation. Our key insight is that model confidence in the decision to engage thinking (or not) can be quantified through entropy analysis of the predicted probability distribution at critical decision tokens. This observation motivates our entropy-governed reward mechanism: the training process naturally transitions from high-entropy exploration, where the model experiments with different thinking strategies, to low-entropy convergence with confident, generalizable decision-making policies. Crucially, this approach enables models to intrinsically discover when to think without requiring manual intervention or external difficulty labels. Extensive experiments demonstrate that our approach enables models to be both accurate on complex problems and efficient on simple ones across diverse document tasks. Code is available at \url{https://github.com/PriNing/AdaThinking-E}.
\end{abstract}    
\section{Introduction}
\label{sec:intro}
Multimodal large language models (MLLMs)~\cite{ChatGPT, GPT-4, GPT-4V} have demonstrated remarkable capabilities in document understanding, ranging from visual text recognition to sophisticated reasoning that solves solid geometry problems, interprets complex layouts, and answers logic-intensive questions. This progress is largely driven by recent advances in reinforcement learning~\cite{jaech2024openai, wei2022chain, shao2024deepseekmathpushinglimitsmathematicalgrpo}, which enable models to generate step-by-step explanations for complex questions.

\begin{figure}[t]
    \centering
    \includegraphics[width=1.\linewidth]{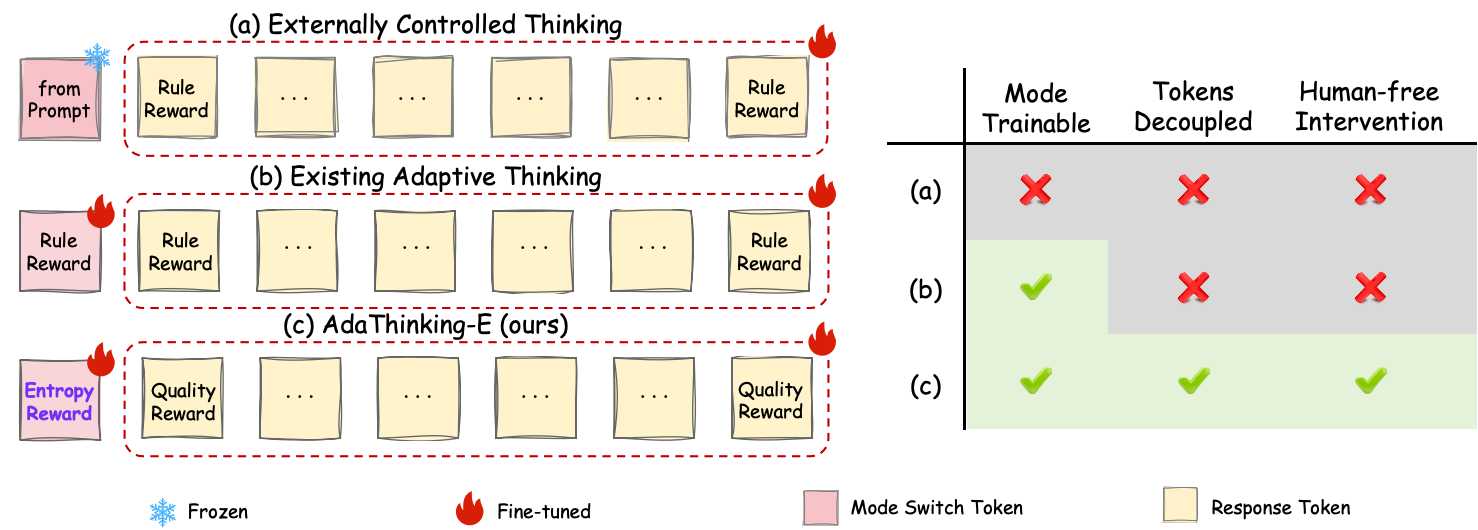}
    \caption{Different RL training paradigms for the mode switch token in adaptive thinking models: (a) \textbf{Externally Controlled Thinking}, the prompt explicitly specifies the output mode (thinking or non-thinking); (b) \textbf{Existing Adaptive Thinking}, the model adaptively switches between thinking and non-thinking modes via special token, using the rule reward values involving subjective human judgments to learn when to think; (c) Our proposed \textbf{AdaThinking-E}, mode switch token and response tokens are optimized separately. Entropy regulation is employed to encourage the model to transition from autonomous exploration to confident decision-making, free from subjective human bias, thereby achieving truly adaptive thinking.}
    \label{fig:abstract}
\end{figure}

However, many everyday question don't require such deep reasoning. Consider a simple question like "Which department issued this memo?"—the answer can be directly extracted from the document header. When applying deep reasoning to all questions, they incur unnecessary computational overhead on simple queries and may even introduce hallucinations through excessive deliberation. Therefore, we argue document MLLMs require an adaptive thinking mechanism~\cite{zhang2505adaptthink, kwaikeyeteam2025kwaikeyevltechnicalreport, yang2025r, tu2025learning} that intelligently determines when to engage thinking based on question complexity. 
This adaptive capability would enable models to be both accurate on complex problems and efficient on simple ones.

Despite its importance, adaptive thinking remains largely unresolved. Current mainstream MLLMs~\cite{xu2025chain, bercovich2025llama, chen2024not, coreteam2025mimovltechnicalreport} rely on manual intervention for mode selection, which lacks flexibility and requires predefined rules that fail to generalize, as illustrated in~\cref{fig:abstract}(a). Alternative approaches~\cite{shen2025dast, li2025selfbudgeter, chen2025overthinker, huang2025adactrl} label question difficulty explicitly, but this introduces subjective judgments, incurs high annotation costs, and ultimately teaches models to follow external labels rather than develop genuine adaptive capabilities, as illustrated in~\cref{fig:abstract}(b).

To deeply analysis the adaptive thinking behavior, we revisit the probability distributions of predicted tokens, where the MLLM estimates the conditional probability of the next token based on previously generated tokens. 
Intuitively, the transition between modes is governed by special tokens within the output. If the model confidently identifies a question as simple, it assigns a higher probability to the token representing the non-thinking mode. In contrast, for complex queries, tokens should have higher probabilities to initiate substantive reasoning. Entropy, which quantifies uncertainty in probability distributions, naturally serves as an indicator of the model's confidence in mode selection.

This insight motivates our formalization of adaptive thinking as an entropy-governed curriculum: transitioning from high entropy states with uncertain mode decisions (exploration) to low entropy states with confident decisions (convergence). During early training, high entropy at mode-switching tokens ensures balanced exploration of both thinking and direct-answer modes. During later training, low entropy ensures consistent, confident mode selection aligned with question complexity.

To achieve this goal, we propose AdaThinking-E, a reinforcement learning framework that learns adaptive thinking through targeted entropy regulation at critical decision tokens, as illustrated in~\cref{fig:abstract}(c). The core innovation lies in independently rewarding entropy dynamics at mode-switching positions: we assign higher advantages to high-entropy outputs in exploration stages and low-entropy outputs in convergence stages, enabling a principled transition from uncertainty to consistency. This mechanism encourages intrinsic self-discovery of when to think rather than relying on externally injected mode identifiers. 
Unlike existing adaptive thinking strategies, we depart from hand-crafted definitions of whether a query necessitates an internal thinking process. Instead, we encourage the model to autonomously explore when to think by regulating the entropy. Furthermore, our decoupled optimization scheme ensures that the actual content of the response focuses exclusively on accuracy, remaining independent of whether the optimal mode selection is achieved.

While our entropy-based framework provides the mechanism for adaptive thinking, effective training requires a dataset that captures the full spectrum of question complexity in document understanding—from simple extraction queries to complex reasoning tasks. To address this gap, we construct AdaThinking-Doc, a document-oriented adaptive thinking dataset designed specifically for training models to develop complexity-aware reasoning capabilities. 

\noindent In summary, our contributions are three-fold:
\begin{itemize}
    \item We propose \textbf{AdaThinking-E}, a novel RL framework that learns adaptive reasoning through entropy-regulated curriculum learning. Unlike prior work requiring manual mode selection or difficulty labels, our approach enables models to intrinsically discover when to think through entropy rewards at decision points.
    \item We introduce \textbf{AdaThinking-Doc}, the first document understanding dataset designed for adaptive thinking. 
    \item Extensive experiments demonstrates that AdaThinking-E achieves state-of-the-art performance across multiple benchmarks, while matching the effectiveness of thinking models with significantly lower token consumption.
\end{itemize}

\section{Related Work}

\subsection{MLLMs for Document Understanding}

MLLMs demonstrate substantial potential in visual document understanding (VDU).
Based on how multimodal features are extracted, these methods can be broadly categorized into OCR-dependent MLLMs~\cite{wang2024docllm, luo2024layoutllm, lu2025bounding, lee2024moai, kim2023visually, tanaka2024instructdoc, liao2025doclayllm,guan2024bridging,guan2023self_,guan2025posformer,guan2023self,guan2022industrial,guan2025ccdplus} and OCR-free MLLMs~\cite{chen2024internvl, zhu2023minigpt, huang2024mini, li2024monkey, hu2025mplug, ye2023ureader, feng2024docpedia, zhang2024token, yu2024texthawk2, shao2024visual, liu2024hrvda, guan2025token, duan2025docopilot, xiao2025adaptive, wang2025marten,guan2026codepercept}. OCR-dependent MLLMs leverage existing OCR models to extract textual and layout information. In contrast, OCR-free MLLMs enable end-to-end VDU by directly processing document images. To strengthen perceptual and reasoning capabilities, researchers have explored both fine-tuning~\cite{li2024monkey, ye2023ureader, hu2025mplug, feng2024docpedia, liu2024hrvda,huang2024mini, shao2024visual, guan2025token, wang2025marten,guo2026vitexqamultiframetemporalperception,jiang2025thinkneedlargehybridreasoning} and reinforcement learning approaches~\cite{yu2025docthinker}.
While prior work~\cite{aggarwal2025l1, luo2025o1, ma2025cot, aytes2025sketch, xu2025chain} has improved reasoning efficiency by compressing output length of reasoning model, these approaches rely on static strategies, limiting their ability to generalize across varying levels of question difficulty. This underscores the need for an adaptive thinking framework capable of employing dynamic reasoning strategies.

\subsection{MLLMs with adaptive thinking}

Reasoning LLMs~\cite{jaech2024openai, wei2022chain} typically generate a chain of thought (CoT) with intermediate steps before arriving at the final answer, offering clear benefits for tasks involving complex computation and logical reasoning. However, excessively long CoT significantly increases inference costs and degrades user experience~\cite{chen2024not, cuadron2025danger}. Recent studies~\cite{zeng2025done, sui2025stop, zhao2025trade, jin2025recut, wu2025more, ghosal2025does} have explored how generation length affects model performance, revealing that the optimal reasoning length varies across tasks. Therefore, the reasoning model should adapt its reasoning depth according to the difficulty of the task.

These methods can be broadly categorized into two types: externally controlled reasoning model~\cite{xu2025chain, bercovich2025llama, chen2024not, coreteam2025mimovltechnicalreport} and adaptive thinking model~\cite{aggarwal2025l1, luo2025o1, lou2025adacot, shen2025dast, li2025selfbudgeter, chen2025overthinker, huang2025adactrl, cheng2025incentivizing, xiang2025just, zhang2505adaptthink, kwaikeyeteam2025kwaikeyevltechnicalreport, yang2025r, tu2025learning}. Externally controlled reasoning model switches between short-form response and long-chain reasoning using external mechanisms such as prompt designs. Llama-Nemotron~\cite{bercovich2025llama} adopts fixed prompt formats like “reasoning on/off” to switch response mode.
In contrast, adaptive thinking models autonomously select the appropriate response mode based on the question, without relying on handcrafted prompt templates. 
Some methods~\cite{shen2025dast, li2025selfbudgeter, chen2025overthinker, huang2025adactrl} explicitly estimate question difficulty or token budget using another model, while others~\cite{cheng2025incentivizing, xiang2025just} use problem-solving rate or observed generation length as implicit indicators of difficulty. More recent work has further investigated models capable of switching between thinking/non-thinking modes, most of them~\cite{zhang2505adaptthink, kwaikeyeteam2025kwaikeyevltechnicalreport, yang2025r} adopt a strategy of explicitly training the model to select between reasoning modes. During the cold-start phase, a dual-mode annealing mechanism equips the model with both reasoning and non-reasoning capabilities, while in the RL stage, carefully crafted reward functions guide the decision of whether to engage in reasoning. 
Heavily influenced by the trainer’s subjective intent, such approaches fail to reflect a genuine understanding and autonomous exploration of mode selection by the adaptive thinking model.
However, existing methods fail to examine the core mechanism of mode switching: the probability distribution when outputting mode-switching tokens. Our method introduces the entropy of the model-switching token into the reward function. By steering entropy variations, it guides the thinking/non-thinking mode switching. 

\section{Methodology}
\begin{figure*}[ht]
    \centering
    \vspace{-3em}
    \includegraphics[width=1.\linewidth]{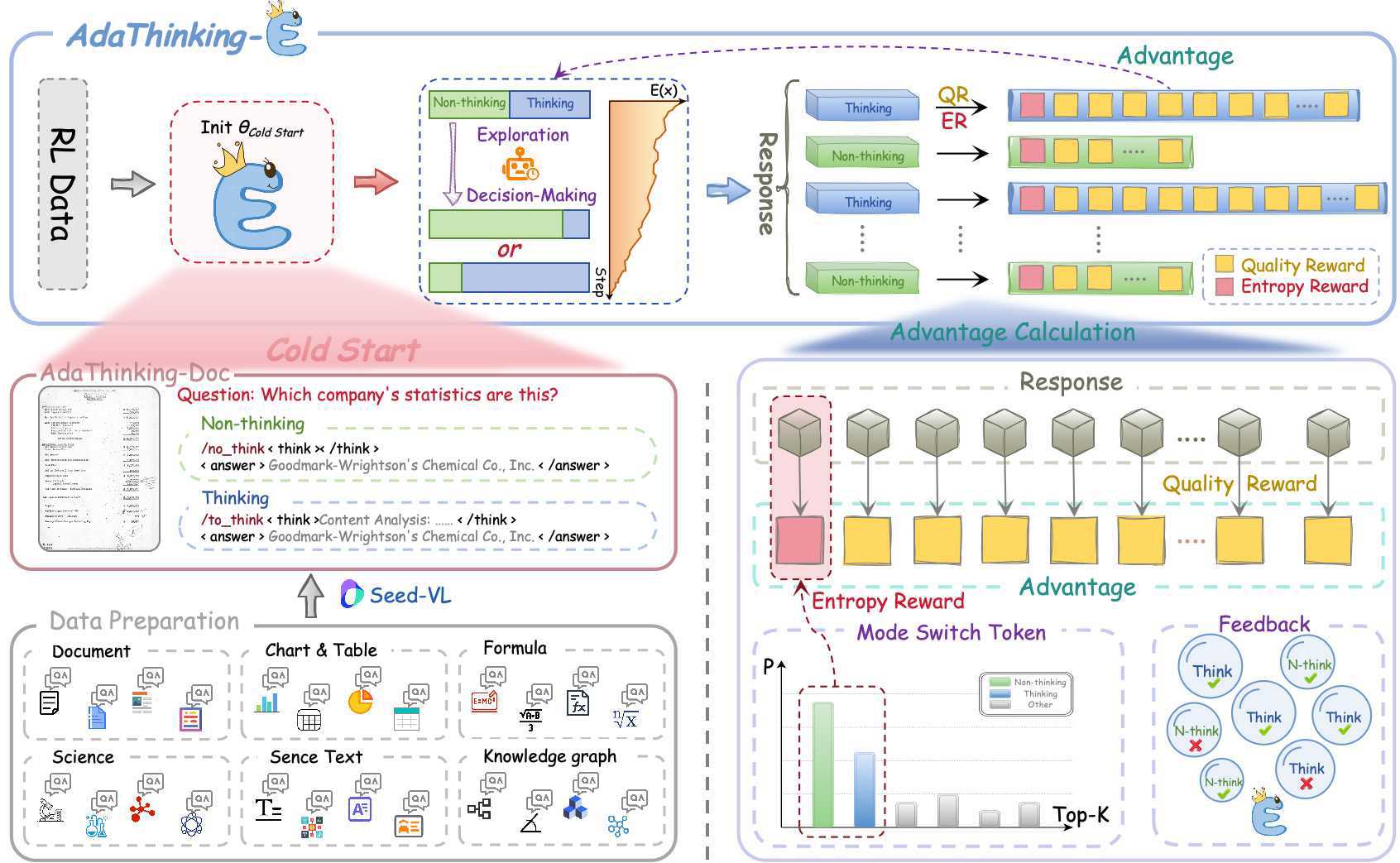}
    \caption{Overview of \textbf{AdaThinking-E}, a reinforcement learning method for adaptive thinking via entropy regulation. The framework has two stages: (1) \textbf{Cold Start}: the model is fine-tuned with AdaThinking-Doc to generate outputs in both thinking and non-thinking modes; (2) \textbf{RL}: the model optimizes two token types separately using cold-start weights as initialization. The mode-switching token transitions from \textbf{high-entropy exploration} to \textbf{low-entropy decision-making}, controlling the sampling ratio between modes. The response tokens (remaining tokens) optimize for answer quality. In the bottom-right, the mode-switching token uses \textbf{Entropy Reward}, while response tokens use content-focused \textbf{Quality Reward}.}
    
    \label{fig:placeholder}
\end{figure*}

In the section, we introduce Adathinking-E, an entropy regulation for refining model behavior, as illustrated in~\cref{fig:placeholder}. The proposed method comprises two distinct training stages: (1) Cold-start stage, where MLLM is fine-tuned to support both concise and detailed response modes, and (2) Reinforcement learning stage, where distinct optimization objectives are designed for different types of tokens. Specifically, the mode switch token regulates the sampling ratio across different modes by controlling the variations in its entropy, thereby learning when to think during this process. Meanwhile, the response tokens (i.e., the remaining tokens) strive to generate correct answers to the maximum extent possible, regardless of the mode selected by the mode switch token.

\subsection{AdaThinking-Doc for Cold-start}
To ensure that our proposed model can stably output both thinking and non-thinking modes in the cold start stage, we constructed AdaThinking-Doc dataset (sourced from existing public datasets) composed of these two modes. For data with thinking mode, we select Seed1.5-VL~\cite{guo2025seed15vltechnicalreport} to generate a detailed reasoning process, which is then validated for logical soundness by GPT-4o-mini~\cite{openai2024gpt4ocard}. 
Regarding the design of the data format, we define the mode switch tokens as \texttt{/no\_think} or \texttt{/to\_think}, representing the non-thinking and thinking modes, respectively.
Meanwhile, the response tokens adhere to the widely adopted format: \texttt{<think>{empty or think content}</think><answer></answer>}.

Additionally, we recognized that manually specifying which questions need thinking, or using model distillation to make this decision, would impose external assumptions onto the model being trained. Such predetermined bias would hinder rather than help the model develop genuine insights during the reinforcement learning (RL) stage. Therefore, to ensure unbiased and random sampling during RL training, we generate both thinking mode and non-thinking mode responses for each data point.

\subsection{Reinforcement Learning}
Following the cold start stage, the model acquires the capability to generate both thinking and non-thinking responses for arbitrary questions, thereby providing sufficiently diverse samples for learning adaptive reasoning abilities during the reinforcement learning stage. To elicit the model's capacity to intelligently determine when to engage deep reasoning based on question complexity, we initially applied GRPO~\cite{shao2024deepseekmathpushinglimitsmathematicalgrpo} algorithm directly. Unfortunately, as training progressed, we observed a systematic collapse: the model rapidly converged to exclusively generating non-thinking responses for all questions, regardless of their complexity. This behavior stems from a fundamental reward attribution problem in standard RL formulations. Specifically, when rewards are distributed uniformly across all tokens in a response, shorter sequences inherently receive higher per-token rewards than longer ones, creating an implicit bias against reasoning-intensive responses. DAPO~\cite{yu2025dapoopensourcellmreinforcementdapo} attempts to mitigate this length bias through reward balancing mechanisms, fail to capture the nuanced decision-making required for adaptive thinking, as they treat all tokens uniformly without considering their distinct roles in mode selection versus content generation.

To gain a deeper understanding of adaptive thinking behaviors, we revisit existing approaches capable of dual-mode output. Through empirical analysis, we observe that the critical determinant for mode selection lies in the heuristic-capable mode switch token, whereas the function of response tokens is merely to provide the answer deemed correct by the model. This observation implies that the decision to initiate reasoning depends exclusively on the mode switch token and is independent of other tokens.

Building on this insight, we propose a novel reward attribution framework that explicitly disentangles the contributions of different token types in the response sequence. Specifically, we categorize tokens into two distinct groups: \textit{mode-switching tokens}, which determine the adaptive thinking mode, and \textit{response tokens}, which constitute the actual reasoning content. 
This decomposition allows us to formulate a dual-objective reward mechanism that optimizes each token type according to its functional role:
\begin{equation}
\small
\begin{aligned}
& \mathcal{J}_{AdaThinking-E} (\theta) = \frac{1}{\sum_{i=1}^{G} |o_i|}
   \sum_{i=1}^{G} \left[
      \mathbb{L}(\theta, E_{i,0}) + \sum_{\substack{t=1}}^{|o_i|} \mathbb{L}(\theta, \hat{A}_{i,t}) 
   \right]
\end{aligned}
\label{eq:jadat}
\end{equation}
where $\theta$ denotes model parameters, $G$ represents the number of samples, $o_i$ represents the $i$-th generated response with length $|o_i|$, and $\mathbb{L}(\theta, \cdot)$ is the loss function, which details refer to supplementary materials (Section 1). Critically, we distinguish between \textit{response tokens} ($t \neq 0$) optimized via advantage estimates $\hat{A}_{i,t}$ and the \textit{mode-switching token} ($t=0$) optimized via $E_{i,0}$ to separately reward content quality and adaptive thinking mode selection.

\subsection{Entropy Reward for Mode-Switching Token}
The entropy magnitude reflects the prediction uncertainty; \emph{e.g.}, higher entropy of the mode-switching token indicates greater decision uncertainty between the two modes. Therefore, we regulate the entropy of the mode-switching token to guide the model's transition from an initial high-entropy exploratory stage to a low-entropy decision-making stage.

\noindent \textbf{Entropy in Mode Switch Token.} Following the standard GRPO algorithm, for each input question $q$ in the training batch, we sample a group of $G$ responses from the current policy $\theta$, where each sample independently draws the mode-switching token according to its predicted probability distribution. This repeated sampling enables us to observe the model's mode selection behavior across multiple trials, thereby providing a robust estimate of its decision entropy. Let $p_{tk}$ and $p_{ntk}$ denote the probabilities of the mode-switching token belonging to thinking mode and non-thinking mode, respectively. For the $i$-th sampled response, the entropy reward of its mode-switching token is defined as:
\begin{equation}
H_{i,0} = - \frac{1}{\ln_{}{2} } \sum_{m \in \{tk, ntk\}} p_m^{(i)}\ln_{}{\left(p_m^{(i)}\right)}
\end{equation}
where $H_{i,0}$ denotes the normalized entropy of the mode-switching token for the $i$-th sample among the $G$ samples generated for the current question, $p_m^{(i)}$ represents the probability of mode $m$ in sample $i$.

We formulate the learning process of adaptive thinking as a two-stage curriculum: (1) High-Entropy Reward for Exploration. In this phase, we aim to maintain a relatively high entropy in mode selection, reflecting uncertainty regarding when to initiate the thinking process. Consequently, the probabilities assigned to the two modes, denoted as  $p_{tk}$ and $p_{ntk}$, should be approximately equivalent. We refrain from imposing manual intervention regardless of the mode selected by the model. This strategy encourages extensive exploration for each query, allowing the model to discover the most suitable mode autonomously. (2) Low-Entropy Reward for Decision-Making. Following the comprehensive evaluation facilitated by the exploration phase, the process gradually transitions towards convergence to reinforce decisive mode selection. In this stage, the model is expected to yield consistent response patterns for specific inputs. Therefore, we incentivize the mode switch token to generate a low-entropy distribution, where the probability of one mode (i.e., $p_{tk}$ or $p_{ntk}$) significantly dominates the other. This shift from high to low entropy indicates that the model has successfully identified the optimal mode for a given problem type, marking a transition from uncertain exploration to confident specialization.

\noindent \textbf{Dynamic Scoring Mechanism.} To facilitate a seamless transition from the exploration phase to the convergence phase, we introduce an adaptive coefficient $\alpha_k$ that varies with the iteration step $k$. This coefficient ensures that high-entropy rewards are amplified during the initial stages. Conversely, as $k$ increases, the gains from high entropy gradually diminish while those from low entropy increase. The specific formulation is as follows:
\begin{equation}
    \alpha_k = \frac{1}{1 + \exp\left[ \beta \times \left( \gamma  - \frac{k}{K} \right) \right]}
\end{equation}
where $K$ represents the total number of steps, $k$ denotes the current step index ($0 \le k \le K$), $\gamma$ signifies the proportion of the total steps at which the inflection point of the transition between two entropy states occurs, and $\beta$ is the smoothness coefficient controlling the steepness of the transition.

\noindent \textbf{Decision Feedback.} 
Besides encouraging the model to explore when to think, we also need to make sure this exploration is reliable. To achieve this, we design a feedback mechanism inspired by confidence-based voting. For each input, we sample multiple outputs under two modes: thinking and non-thinking. We then evaluate which mode tends to give more correct answers, and use this as training feedback. This enables the model learn when thinking is useful by itself, rather than relying on predefined human instructions regarding whether to invoke the thinking process, thus fully unlocking its potential for autonomous exploration. Specifically, for each sampled rollout group, we label every rollout by (1) its mode (thinking/non-thinking) and (2) whether its final answer is correct. From this, we compute for each mode \(m \in \{tk, ntk\}\):  

\noindent - \(r_m\): how often this mode appears in the sampled group (occurrence ratio);

\noindent - \(acc_m\): accuracy of this mode within the group.  

Subsequently, we divide accuracy into three levels using thresholds \(\epsilon_l\) and \(\epsilon_h\): low, medium, and high. 
When $acc_m$ in both modes falls within the same tier, an accuracy guidance term $\mathbb{A}$ is triggered to regulate the optimization. Specifically, when accuracy is low in both modes, it guides the model to invoke reasoning; when both $acc_{tk}$ and $acc_{ntk}$ are medium-tier or both are high-tier, it guides the model to bypass reasoning.

Furthermore, we observe that when the model demonstrates a clear propensity in responding to a sample (defined as $|r_{tk} - r_{ntk}| > \Delta_r$), imbalanced sampling may lead to unreliable comparisons that compromise training stability. For instance, consider a sampling instance where the model achieves 13/15 correct answers ($acc_{tk}=86.7\%$) in thinking mode versus 1/1 ($acc_{ntk}=100\%$) in non-thinking mode. Naive $\mathcal{R}_{Ada}$ would steer optimization toward the non-thinking direction, contradicting model's demonstrated correct decisions, which results in training instability or even optimizing collapse. Therefore, we introduce a propensity guidance term $\mathbb{P}$ which guides the model towards its propensity when it demonstrates a clear propensity and both modes achieve high accuracy.

Finally, the weight for mode \(m\) is computed as:
\[
w_m = \frac{acc_m}{acc_{\neg m}} + \mathbb{A}(acc, m) + \mathbb{P}(acc, r, m,\Delta_r)
\]

More details about each term and its value in different cases are provided in Supplementary Materials (Section 2).

\noindent \textbf{Entropy Reward.} Finally, the overall entropy reward function is formulated in the following:
\begin{equation}
\small
E_{i,0} = 
      w_m\left[\underbrace{\left (1 - \alpha_k\right )H_{i,0}}_{high-entropy} + \underbrace{\alpha_k\left (1-H_{i,0}\right )}_{low-entropy}\right ]+ p_{tk} + p_{ntk}
\end{equation}
where the additive term $p_t + p_{nt}$ ensures that the cumulative probability of sampling a mode switch token at the first position approaches 1, thereby precluding the possibility of the first token being any other alternative identifier.

\subsection{Quality Reward for Response Tokens}
While entropy regulation governs the mode-switching decision, it remains agnostic to the quality of reasoning and problem-solving within each mode. Specifically, the entropy reward only determines \textit{whether} thinking mode is activated, but cannot assess \textit{how well} the model reasons in thinking mode or \textit{how accurately} it responds in non-thinking mode. To address this, we introduce a content-focused reward function that directly evaluates the problem-solving effectiveness of the response tokens, thereby ensuring that appropriate mode selection is complemented by high-quality content generation.

\noindent \textbf{Format Reward.} The format reward $R_f$ ensures strict adherence to the predefined template structure. This is critical for our entropy-based mode-switching mechanism that relies on specific token positions. We define \texttt{isValidFormat}(S) to verify template compliance: 
\begin{align} 
R_f = \begin{cases} 1, & \text{if isValidFormat}(S), \\ 0, & \text{otherwise.} \end{cases} 
\end{align}

\noindent \textbf{Accuracy Reward.} The accuracy reward $R_{acc}$ evaluates whether the final answer $\hat{y}$ matches the ground truth $y$, independent of the reasoning mode: 
\begin{equation}
    R_{acc} = \left\{\begin{matrix}
 1, & \text{if} \; \hat{y} = y, \\
 0, & \text{otherwise.}
\end{matrix}\right.
\end{equation}

\section{Experiments}

\subsection{Training Details}
\label{sec:41}
\noindent \textbf{Datasets.} All data utilized by AdaThinking-Doc are sourced from the training sets of publicly available datasets, encompassing both simple information extraction and complex document reasoning scenarios. Detailed information regarding the data distribution is provided in the Supplementary Material.

\noindent \textbf{Training Details.} We used QwenVL-2.5-7B~\cite{bai2025qwen25vltechnicalreport} or QwenVL-3-8B~\cite{bai2025qwen3vltechnicalreport} as the base model. For the Cold-start stage, we fine-tuned using SWIFT~\cite{zhao2024swiftascalablelightweightinfrastructure} with learning rate 1e-6, 1 epoch, and batch size 4. We then conducted RL training using VeRL~\cite{sheng2024hybridflow} with learning rate 1e-6, batch size 32, 16 samples per prompt, and 1 epoch for 2000 steps. All experiments used 8 NVIDIA A100 GPUs.

\noindent \textbf{Evaluation.} We evaluated AdaThinking-E against existing MLLMs, including non-thinking (NT), thinking (TK), and adaptive thinking (ATK) models, using VLMEvalKit~\cite{duan2024vlmevalkit} across multiple benchmarks~\cite{mathew2021docvqa,mathew2022infographicvqa,singh2019towardstextvqa,masry2022chartqa,wang2024charxiv,liu2023hiddenocrbench,huang2025ocr}. GPT-4o-mini~\cite{openai2024gpt4ocard} served as the evaluator for tasks requiring large model assessment. All ablation studies are conducted using Qwen2.5-VL-7B~\cite{bai2025qwen25vltechnicalreport} as the base model.

\subsection{Main Results}

\begin{table*}[ht]
    \centering
    \caption{Performance comparison of MLLM on on diverse document benchmarks. "NT", "TK", and "ATK" denote the Non-Thinking, Thinking, and Adaptive Thinking modes, respectively. "*" indicates that the corresponding mode switch token is proactively inserted at the input stage according to the selected mode during the inference phase."$\dag$" denotes results reproduced from the original implementation.}
    \scalebox{0.6}{
    \begin{tabular}{c|c|ccc|ccc|cc}
        \toprule 
        \textbf{Model} & \textbf{Mode} & \textbf{DocVQA} & \textbf{InfoVQA} & \textbf{Text\textsubscript{Val}} & \textbf{ChartQA} & \textbf{CharXiv$_{RQ}$} & \textbf{CharXiv$_{DQ}$} & \textbf{OCR-Bench} & \textbf{OCR-Reasoning}  \\
        \midrule
        DocOwl-1.5-8B~\cite{hu2024mplug} & NT & 81.6 & 50.4 & 68.8 & 70.5 & - & - & 59.9 & -  \\
        Monkey-10B~\cite{li2024monkey} & NT & 66.5 & 36.1 & 67.6 & 65.1 & - & - & 51.4 & -  \\
        TextMonkey-8B~\cite{liu2024textmonkey} & NT & 73.0 & 28.6 & 65.6 & 66.9 & - & - & 56.1 & -  \\
        MiniMonkey-2B~\cite{huang2024mini} & NT & 87.4 & 60.1 & 75.7 & 76.5 & - & - & 80.2 & -  \\
        HRVDA-7B~\cite{liu2024hrvda} & NT & 72.1 & 43.5 & 77.3 & 67.6 & - & - & - & - \\
        InternVL2.5-7B~\cite{chen2025expandingperformanceboundariesopensource} & NT & 93.0 & 77.6 & 79.1 & 84.8 & 32.9 & 68.6 & 82.2 & - \\
        InternVL3-8B~\cite{zhu2025internvl3exploringadvancedtraining} & NT & 92.7 & 76.8 & 80.2 & 86.6 & 37.6 & 73.6 & 88.0 & 11.5  \\
        QwenVL2.5-8B~\cite{bai2025qwen25vltechnicalreport} & NT & 95.7 & 82.6 & \underline{84.9} & 87.3 & 42.5 & 73.9 & 86.4 & 15.7  \\
        QwenVL3-8B~\cite{bai2025qwen3vltechnicalreport} & NT & 96.1 & 83.1 & 81.9 & 89.6 & 46.4 & 83.0 & 89.6 & 18.5  \\
        TextHawk2-7B~\cite{yu2024texthawk2} & NT & 89.6 & 67.8 & 75.1 & 81.4 & - & - & 78.4 & -  \\
        Marten-7B\cite{wang2025marten} & NT & 92.0 & 75.2 & 74.4 & 81.7 & - & - & 82.0 & -  \\
        AlignVLM-8B~\cite{masry2025alignvlmbridgingvisionlanguage} & NT & 81.2 & 53.8 & 64.6 & 75.0 & - & - & - & - \\
        TokenVL-7B~\cite{guan2025token} & NT & 94.2 & 76.5 & 79.9 & 86.6 & - & - & 86.0 & 14.3 \\
        DocMark-2B~\cite{xiao2025adaptive} & NT & - & - & 74.8 & 79.8 & - & - & 81.3 & 7.4 \\
        Docopilot-8B~\cite{duan2025docopilot} & NT & 92.0 & 73.3 & - & 83.3 & - & - & - & 11.6 \\
        \hline
        OpenVLThinker-7B$^\dag$~\cite{deng2025openvlthinker} & TK & 94.6 & 76.5 & 81.8 & 82.7 & 42.4 & 66.2 & 81.4 & 23.3 \\
        R1-Onevision-7B$^\dag$~\cite{yang2025r1}  & TK & 92.4 & 76.6 & 75.4 & 82.3 & 37.1 & 54.9 & - & 21.2  \\
        VLAA-Thinker-7B$^\dag$~\cite{chen2025sftrlearlyinvestigation} & TK & 95.7 & 80.1 & 82.9 & 87.3 & 38.9 & 73.3 & 83.0 & 14.4 \\
        Kimi-VL-A3B-Thinking~\cite{kimiteam2025kimivltechnicalreport} & TK & - & - & - & - & 47.7 & 75.4 & 82.5 & 20.5 \\
        VL-Rethinker-7B$^\dag$~\cite{wang2025vl} & TK & 95.9 & 78.9 & 82.7 & 87.1 & 43.3 & 64.9 & 86.4 & 14.6  \\
        DocThinker-7B~\cite{yu2025docthinker} & TK & - & - & 83.6 & - & - & - & - & -  \\
        QwenVL3-8B~\cite{bai2025qwen3vltechnicalreport} & TK & 95.3 & 86.0 & 78.7 & 88.6 & 53.0 & 85.9 & 81.9 & 48.4  \\
        \hline
        R-4B$^\dag$~\cite{yang2025r4bincentivizinggeneralpurposeautothinking} & ATK & 94.1 & 67.0 & 76.6 & 87.2 & 56.8 & 82.9 & 83.6 & 22.2  \\
        Keye-VL-8B$^\dag$~\cite{kwaikeyeteam2025kwaikeyevltechnicalreport} & ATK & 90.5 & 62.6 & 78.6 & 82.4 & 40.0 & 74.5 & 85.3 & 22.6  \\
        Mimo-VL-7B$^\dag$~\cite{coreteam2025mimovltechnicalreport} & ATK & 96.1 & 79.7 & 80.6 & 87.4 & 56.5 & \textbf{86.8} & 86.6 & -  \\
        ARES-7B$^\dag$~\cite{chen2025ares} & ATK & 92.1 & 72.2 & 80.5 & 88.2 & 51.0 & 79.3 & 83.1 & 23.4  \\
        \hline
        \rowcolor{lightgray!40}
        AdaThink-E-7B-Qwen2.5* & NT & \underline{96.3} & 82.8 & \underline{84.9} & 88.4 & 55.2 & 82.6 & 86.9 & 21.7   \\
        \rowcolor{lightgray!40}
        AdaThink-E-7B-Qwen2.5* & TK & 95.1 & 80.8 & 84.5 & 88.5 & 56.2 & 83.3 & 87.3 & 26.1 \\
        \rowcolor{lightgray!70}
        AdaThink-E-7B-Qwen2.5 & ATK & \underline{96.3} & 82.9 & \textbf{85.3} & 89.1 & 56.7 & 83.5 & 88.7 & 26.5  \\
        \hline
        \rowcolor{lightgray!40}
        AdaThink-E-8B-Qwen3* & NT &  96.2 & 84.7 & 82.6 & 89.3 & 54.2 & 83.9 & \underline{88.9} & 31.2 \\
        \rowcolor{lightgray!40}
        AdaThink-E-8B-Qwen3* & TK & 95.7 & \underline{86.2} & 80.9 & \underline{89.9} & \underline{56.9} & 86.3 & 87.7 & \underline{49.4} \\
        \rowcolor{lightgray!70}
        AdaThink-E-8B-Qwen3 & ATK & \textbf{96.4} & \textbf{86.3} & 82.8 & \textbf{90.1} & \textbf{57.3} & \underline{86.6} & \textbf{89.8} & \textbf{49.7} \\
        \bottomrule
    \end{tabular}
    }
    \label{tab:res1}
\end{table*}

\begin{figure*}[htbp]
    \centering
    \vspace{-1em}
    \includegraphics[width=1.\linewidth]{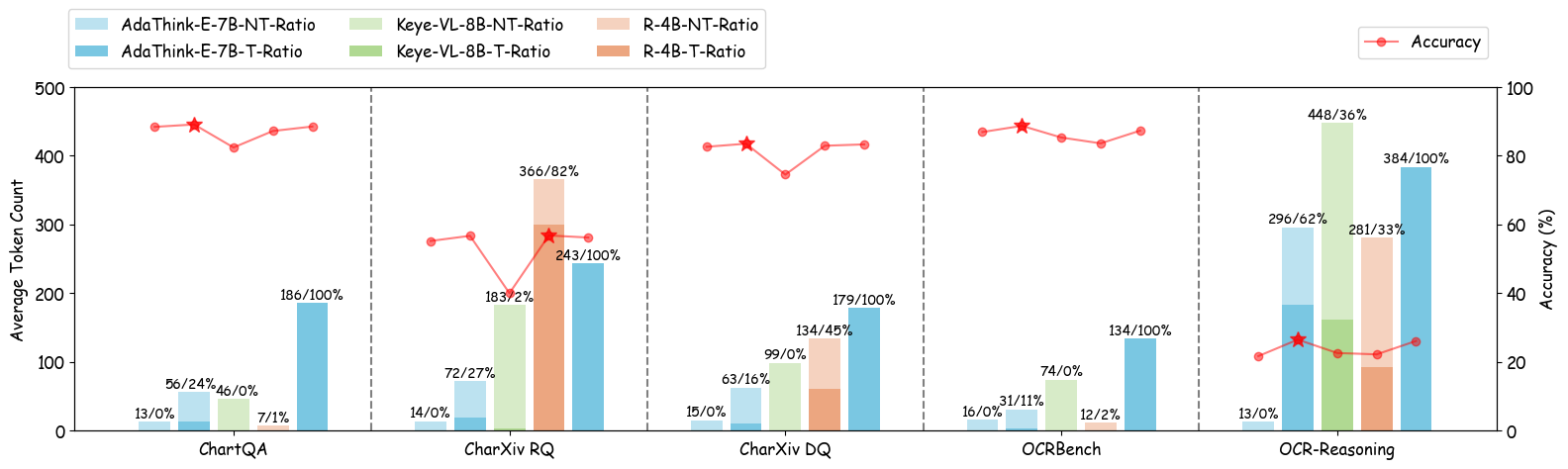}
    \caption{Comparison of multiple metrics for the adaptive thinking model across document understanding benchmarks. Results of AdaThinking-E include non-thinking, adaptive thinking, and thinking modes.  Bar height represents average output tokens, internal color ratio indicates thinking ratio, and the line plot shows accuracy. In the blue bars, 0\% and 100\% thinking ratios correspond to the non-thinking and thinking modes of AdaThinking-E, respectively.}
    \label{fig:RES}
\end{figure*}

\noindent \textbf{VQA-based Document Understanding.} As shown in~\cref{tab:res1}, AdaThinking-E is compared against three types of MLLMs with distinct reasoning modes across multiple document understanding benchmarks, consistently achieving superior performance. On chart-based datasets requiring analytical computation, such as CharXiv, models equipped with a thinking mode (TK) outperform those with a Non-Thinking mode. However, in information extraction scenarios, the advantage of the thinking mode becomes less pronounced and can even lead to performance degradation due to overthinking-induced hallucinations. 

Our proposed AdaThinking-E achieves near SOTA performance across both information extraction and analytical computation tasks. AdaThinking-E’s adaptive thinking surpasses the baseline Qwen2.5-VL by 0.6\% on DocVQA, and achieves improvements of 1.8\% and 10.8\% on ChartQA and OCR-Reasoning, respectively. Even compared to R-4B, which also features adaptive thinking, AdaThinking-E-Qwen2.5 delivers gains of 1.9\% and 4.3\% on these two datasets. These results clearly demonstrate AdaThinking-E-Qwen2.5’s superior adaptive thinking capabilities in document understanding tasks. AdaThinking-E-Qwen3, trained on Qwen3-VL-Thinking, outperforms Qwen3-VL-Thinking across various benchmarks and even surpasses the higher-performing Qwen3-VL-Instruct on certain tasks. Additionally, in the last six rows of~\cref{tab:res1}, we compare the performance of AdaThinking-E under different modes. The thinking and non-Thinking modes are actively switched via mode switch token. It is evident that the adaptive thinking mode consistently matches or exceeds the performance of each standalone mode across various benchmarks, further validating the adaptive thinking capabilities of AdaThinking-E.

\begin{figure*}
    \centering
    \vspace{-1.em}
    \includegraphics[width=1.\linewidth]{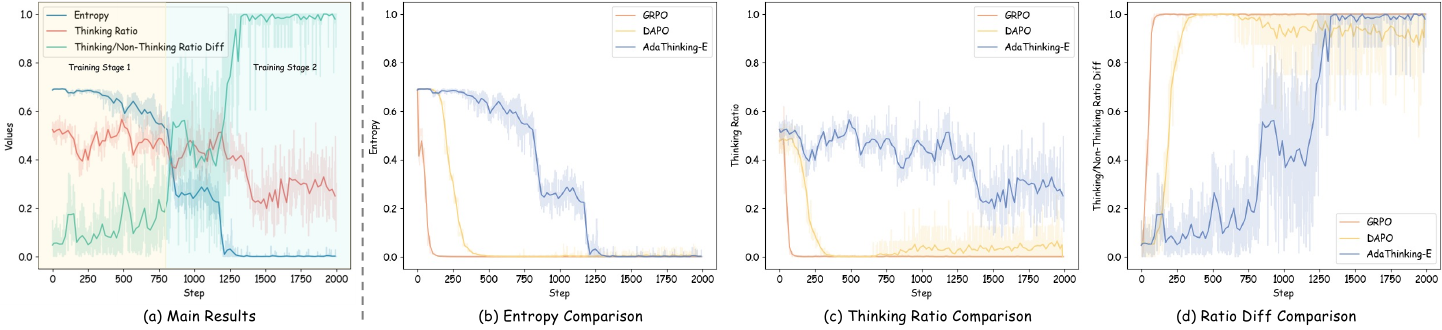}
    \caption{Illustrate the trends in entropy and thinking ratio changes during the training phase. (a) displays the variation curves of three metrics of AdaThinking-E as the training steps increase. Specifically, "Entropy" represents the average entropy of mode switch tokens within a step, "Thinking Ratio" indicates the proportion of thinking samples within a step, and "Thinking/Non-Thinking Ratio" denotes the average difference in the ratio of thinking and non-thinking modes within the group. (b), (c), and (d) showcase the comparison of the three metrics between AdaThinking-E, GRPO, and DAPO under the same reward conditions}
    \label{fig:exp}
    \vspace{-.5em}
\end{figure*}

\noindent \textbf{Token Count and Thinking Ratio.} As illustrated in~\cref{fig:RES}, we analyzed the average output length, thinking ratio, and accuracy of each query under different modes in AdaThinking-E-Qwen2.5, and compared these metrics with advanced adaptive thinking MLLMs. 
To ensure fairness in output token length, we uniformly added the phrase "Answer the question using a single word or phrase." in each query, ensuring that the token length of the answer part remains relatively balanced apart from the thinking content. On the CharXiv$_{RQ}$ dataset, AdaThinking-E achieved a comparable level to R-4B with only a 27\% thinking ratio (72 tokens) compared to R-4B's 82\% thinking rate (366 tokens). Moreover, on CharXiv$_{DQ}$, AdaThinking-E outperformed R-4B by 0.6\% in performance with a lower thinking rate (11\% compared to R-4B's 45\%). This demonstrates that AdaThinking-E exhibits superior document understanding capabilities compared to other adaptive thinking models.Additionally, on the OCR-Reasoning benchmark, AdaThinking-E's thinking ratio increased to 62\%, with an accuracy rate of 26.5\%—0.4\% higher than its own thinking mode's 26.1\%—while the average token count decreased by 88. In contrast, R-4B and Keye-VL showed only 33\% and 38\% thinking ratios, respectively, and their performance decreased by 4.3\% and 3.9\% compared to AdaThink-E due to excessive non-thinking decisions. These findings collectively prove AdaThinking-E has a more accurate understanding of document complexity, maintaining performance while ensuring efficiency, and shows robust capabilities.

\subsection{Training Analyze}
To further analyze how entropy influences the learning of mode-switching strategies in adaptive thinking models, in~\cref{fig:exp}, we visualized the indicators related to entropy and thinking ratio during the RL stage. \cref{fig:exp}(c) illustrates the changes in thinking ratios of three strategies. Both GRPO and DAPO quickly tend towards the non-thinking mode during training, resulting in mode collapse. 

\cref{fig:exp}(a) demonstrates how AdaThinking-E effectively addresses the mode collapse issue by regulating entropy during training. AdaThinking-E decouples mode switch token from response tokens, linking the gradient of mode switch token solely to entropy levels, preventing entropy rewards from affecting response tokens. During the exploratory stage, entropy remains at a high level, ensuring a balanced sampling with low differences in the number of thinking and non-thinking modes within the group, which provides high-quality samples for optimizing mode-switching strategies. As training progresses into the decision-making stage, mode collapse does not occur, and the proportion of thinking and non-thinking within the step remains relatively stable.

\subsection{Ablation Study}

\noindent \textbf{Hyperparameter Setting in Decision Feedback.} We conduct comprehensive ablation studies on three key hyperparameters in decision feedback. As shown in~\cref{tab:abs_adr}, two benchmarks of complementary difficulty levels, ChartQA (easier) and OCR-Reasoning (more challenging) are selected to evaluate scenario specific impacts. 
In simple scenarios, a lower $\epsilon_h$ increases the activation probability of $\mathbb{A}$, thereby amplifying the influence of $\mathbb{P}$. Higher $\Delta_r$ impose stricter criteria for determining model propensity, which impedes activation of $\mathbb{P}$, resulting in a pronounced reduction in the model's reasoning frequency. In challenging scenarios, the model's lower accuracy impedes activation of $\mathbb{P}$, while higher $\epsilon_l$ settings facilitate triggering of $\mathbb{A}$, enhancing both thinking frequency and accuracy. Furthermore, we investigate the performance when relying solely on $acc$ as decision feedback, omitting $\mathbb{P}$ and $\mathbb{A}$. While this configuration achieves comparable scores in terms of performance, its thinking ratio is significantly higher than the results obtained with $\mathbb{P}$ and $\mathbb{A}$. This suggests that $\mathbb{P}$ and $\mathbb{A}$ effectively facilitate a deeper exploration of the transition from thinking to non-thinking modes.

\begin{table}[htbp]
    \centering
    
    \caption{Hyperparameter Ablation in Decision Feedback. "Think (\%)" represents the ratio of instances where the thinking mode is activated. "$\beta$" serves as the smoothing coefficient. "-/-" indicates that the parameter is not used.}
    \label{tab:abs_adr}
    \setlength{\tabcolsep}{20pt} 
    \resizebox{1.\textwidth}{!}{
    \begin{tabular}{cccccccc}
    \toprule
    \multirow{2}{*}{method} & \multirow{2}{*}{$\epsilon_l$} & \multirow{2}{*}{$\epsilon_h$} & \multirow{2}{*}{$\Delta_r$} & \multicolumn{2}{c}{\textbf{ChartQA}} & \multicolumn{2}{c}{\textbf{OCR-Reasoning}} \\
    \cmidrule(lr){5-6}
    \cmidrule(lr){7-8}
     &  &  & & Acc & Think(\%) & Acc & Think(\%) \\
    \midrule
    w/o $\mathbb{A}$\&$\mathbb{P}$ & -/- & -/- & -/- & 88.1 & 39.9 & 25.7 & 83.4 \\
    \midrule
      \multirow{3}{*}{w $\mathbb{A}$\&$\mathbb{P}$} & \multirow{3}{*}{0.2} & \multirow{3}{*}{0.8} & 0.625 & \underline{88.6} & 8.7 & \underline{25.6} & 55.1 \\
     & & & 0.5 & 88.5 & 12.5 & 25.5 & 54.7 \\
     & & & 0.375 & \textbf{89.0} & 23.7 & \textbf{25.8} & 56.6 \\
     \midrule
     \multirow{3}{*}{w $\mathbb{A}$\&$\mathbb{P}$} & \multirow{3}{*}{0.3} & \multirow{3}{*}{0.7} & 0.625 & 88.4 & 3.4 & 25.9 & 59.7 \\
     & &  & 0.5 & \underline{88.9} & 13.1 & \underline{26.2} & 60.3 \\
     & &  & 0.375 & \textbf{89.1} & 24.3 & \textbf{26.5} & 61.9 \\
    \bottomrule
    \end{tabular}}
\end{table}

\noindent \textbf{Entropy Regulation.} \cref{tab:abs_entroy} compares the performance and thinking ratio on document benchmarks with various parameter configurations. 
The first row of the table represents the cold-start baseline. Compared with Qwen2.5-VL, the model fine-tuned using AdaThinking-Doc during the cold-start phase achieves performance gains of 0.6\% on ChartQA and 5.9\% on OCR-Reasoning, respectively, demonstrating its effectiveness.
In rows 2–3, we simulate scenarios that encourage either high or low entropy throughout the entire process. It can be observed that maintaining consistently high entropy for mode switch tokens leads to unstable thinking ratios and misallocates challenging queries to the non-thinking mode, which results in a performance drop of approximately 1.5\%. Conversely, in low-entropy states, the mode-switching tokens gain an overwhelming bias toward non-thinking patterns, causing the model to converge prematurely to a non-thinking state. This leads to mode collapse and a significant reduction in the overall score by about 5\%. 

We explore the configurations of the inflection point $\gamma$ and the smoothing coefficient $\beta$ during the exploration and convergence phases. When $\gamma$ is larger, the model accuracy is nearly consistent with that when $\gamma$ is smaller, but the overall thinking ratio is 18\% higher. This indicates that during the exploratory stage, due to balanced sampling, the progress of exploring the non-thinking mode is slowed. An excessively small smoothing coefficient renders the model insensitive to entropy regulation, leading to suboptimal performance. For a more comprehensive analysis, please refer to the Supplementary Material.

\begin{table}[htbp]
    \centering
    \caption{Performance comparison of entropy regulation under different parameter configurations. "$\gamma$" represents the inflection point of the transitionand, and "$\beta$" serves as the smoothing coefficient. "-/-" indicates that the parameter is not used.}
    \label{tab:abs_entroy}
    \setlength{\tabcolsep}{20pt} 
    \resizebox{1.\textwidth}{!}{
    \begin{tabular}{ccccccc}
    \toprule
     \multirow{2}{*}{stage} & \multirow{2}{*}{$\beta$} & \multirow{2}{*}{$\gamma$}  & \multicolumn{2}{c}{\textbf{ChartQA}} & \multicolumn{2}{c}{\textbf{OCR-Reasoning}} \\
     \cmidrule(lr){4-5}
     \cmidrule(lr){6-7}
     & & & Acc(\%) & Think(\%) & Acc(\%) & Think(\%) \\
    \midrule
    Cold-start & -/- & -/- & 87.6 & 9.2 & 21.6 & 10.4 \\
    \midrule
      \multirow{2}{*}{RL} & \multirow{2}{*}{-/-} & 1.0 & 88.1 & 75.6 & 23.7 & 48.4 \\
    & & 0.0  & 87.8 & 0.0 & 18.7 & 0.0 \\
    \midrule
    \multirow{3}{*}{RL} & \multirow{3}{*}{5} & 0.6 & \underline{88.4} & 50.7 & 25.5 & 63.3 \\
    & & 0.5 & 88.3 & 48.3 & \underline{25.7} & 67.2 \\
    & & 0.4 & \textbf{88.6} & 40.6 & \textbf{26.1} & 59.7 \\
    \midrule
     \multirow{3}{*}{RL} & \multirow{3}{*}{10} & 0.6 & \underline{88.9} & 39.3 & \underline{26.7} & 83.8 \\
     & & 0.5 & 88.8 & 30.6 & \textbf{26.9} & 75.6 \\
     & & 0.4 & \textbf{89.1} & 24.3 & 26.5 & 61.9 \\
    \bottomrule
    \end{tabular}
    }
\end{table}
\section{Conclusion}
In this work, we introduce \textbf{AdaThinking-E}, a reinforcement learning framework that enables multimodal large language models to dynamically switch between thinking and non-thinking modes according to task complexity. By leveraging one-token entropy regulation, our method  incorporates an entropy-based reward to guide the transition from exploratory to convergent thinking strategies. This adaptive thinking allows the model to intrinsically determine when to think without relying on external difficulty annotations, thereby improving efficiency on simple tasks while preserving accuracy on complex ones. Extensive experiments across diverse document understanding benchmarks validate the effectiveness of our approach, demonstrating that AdaThinking-E achieves a favorable balance between computational efficiency and reasoning performance.


\section*{Acknowledgements}
This work was supported by NSFC 62322604 and NSFC 62576207.

%
%

\clearpage

\bibliographystyle{splncs04}
\bibliography{main}

\clearpage
\clearpage
\setcounter{page}{1}
\setcounter{section}{0}
\setcounter{figure}{0}
\setcounter{table}{0}
\setcounter{equation}{0}

\appendix
\title{AdaThinking-\includegraphics[height=1.5em]{fig/E.png}: One-Token Entropy Regulation for Adaptive Thinking \\ (Supplementary Material)} 



\section{Dual-objective Reward Mechanism}
\noindent \textbf{Review of DAPO.} During \textbf{R}einforcement \textbf{L}earning(RL) training stage of \textbf{AdaThinking-E}, we adopt \textbf{DAPO}~\cite{yu2025dapoopensourcellmreinforcementdapo} as the underlying policy optimization method. The formulation of DAPO is defined as:
\begin{equation}
   \mathcal{J}_{DAPO} (\theta) = \frac{1}{\sum_{i=1}^{G} |o_i|}
   \sum_{i=1}^{G} 
      \sum_{\substack{t=1}}^{|o_i|} \mathbb{L}_{i,t}(\theta, \hat{A}_{i,t}) 
\end{equation}
where $\mathbb{L}_{i,t}(\theta, \hat{A}_{i,t})$ is defined as follows:
\begin{equation}
\begin{aligned}
   & \mathbb{L}_{i,t}(\theta, \hat{A}_{i,t}) = \\
   & \textrm{min}\left ( r_{i,t}(\theta )\hat{A}_{i,t}, \textrm{clip}\left (r_{i,t}(\theta ), 1-\varepsilon _{l}, 1+\varepsilon _{h} \right ) \hat{A}_{i,t}\right )
\end{aligned}
\end{equation}
All algorithms are derived directly from DAPO. Compared to \textbf{GRPO}~\cite{shao2024deepseekmathpushinglimitsmathematicalgrpo}, DAPO introduces the following modifications: (1) \textbf{Token-level policy gradient loss:} DAPO adopts token-level loss aggregation, averaging over all tokens in a batch, replacing the sequence-level averaging approach used by GRPO. This ensures that each token's contribution to the gradient update is equal, regardless of whether it comes from a long or short sequence, thus avoiding the dilution of token contributions from longer sequences. (2) \textbf{Higher truncation:} DAPO modifies the truncation mechanism of GRPO by decoupling the upper $\varepsilon _{h}$ and lower $\varepsilon _{l}$ bounds of the truncation interval. By setting an upper bound larger than the lower bound, this strategy allows tokens with initially low probabilities more room to increase their probabilities, thereby enhancing the policy entropy and promoting diversity and exploration in generated samples. This prevents the model from becoming deterministic too early, which would limit exploration and lead to local optima. (3) \textbf{Dynamic sampling:} DAPO actively oversamples and discards groups where all responses are either correct or incorrect. This ensures that each batch used for gradient computation contains samples with a mix of correct and incorrect responses, thereby ensuring the presence of non-zero advantage and enabling each sample to provide effective learning signals. (4) \textbf{Removal of KL divergence:} This ensures better exploration during the training of long CoT reasoning models.

DAPO retains the original importance sampling and advantage calculation methods of GRPO. For a given query $q$, the old policy model first generates a set of response samples $\left \{ o_0,o_1,...o_G \right \}$, totaling $G$ samples. These responses are evaluated through a pre-designed reward function to obtain reward values $\left \{ R_0,R_1,...R_G \right \}$, such as the \textbf{Quality Reward Function} proposed in Adathinking-E. Subsequently, the reward values are normalized within the group to obtain relative advantages $\hat{A}_{i,t}$. The formula is as follows:
\begin{equation}
    \hat{A}_{i,t}= \frac{R_i - \textrm{mean}\left ( \left \{ R_i\right \}^G_{i=1}\right )}{\textrm{std} \left( \left \{ R_i\right \}^G_{i=1} \right)}
\end{equation}
where mean(·) and std(·) are the mean and standard deviation of the rewards across the group of responses.

The importance sampling ratio $r_{i,t}$ is used to adjust the weight coefficients of sample advantages, typically applied in the transition from one probability distribution (old policy $\pi_{\theta_{old}}$) to another (new policy $\pi_\theta$). It can correct the estimation of expected returns, making policy updates more stable and effective. The formula is expressed as follows:
\begin{equation}
    r_{i,t}(\theta) = \frac{\pi_\theta (o_{i,t}|q,o_{i<t})}{\pi_{\theta_{old}} (o_{i,t}|q,o_{i<t})} 
\end{equation}
\noindent \textbf{Modifications to DAPO.} For $\mathcal{J}_{AdaThinking-E}$, the modification is applied to $\hat{A}_{i,3}$, which is replaced by $E_{i,3}$ as defined in Eq. (4) from Sec. 3.3 of the main text. Additionally, AdaThinking-E has modified the Dynamic sampling of DAPO by retaining the groups with correct responses. This consideration stems from our fundamental goal of guiding the model to learn when to think. The sampling ratio of thinking and Non-Thinking within the groups with correct responses can still provide effective learning signals during model training.

\section{Decision Feedback}

Unlike traditional methods, \textbf{Decision Feedback} transcends the limitations of single sampling by providing performance feedback for each query based on a collective set of samples.

\noindent \textbf{Accuracy and Occurrence Ratio.} The core design is derived from two parameters: the occurrence ratio $\rho_k$ and accuracy $acc_k$ for thinking $tk$ and non-thinking $ntk$ modes($k \in \{tk, ntk\}$), which are represented as follows:
\begin{equation}
    \rho_k = \frac{\left | \left \{  o_i| \hat{k_{i}} = k \right \}  \right | }{G}
\end{equation}
\begin{equation}
     acc_k = clip \left ( \frac{\left | \left \{  o_i| \hat{y_{i}} = y, \hat{k_{i}} = k \right \}  \right | }{\left | \left \{  o_i| \hat{k_{i}} = k \right \}  \right | + 10^{-8}} , \epsilon_l, \epsilon_h \right )
\end{equation}
where $\hat{k_{i}}$ denotes the mode to which the $i$-th sample belongs, $y$ denotes the ground truth of the query, and $\hat{y_{i}}$ denotes the prediction of the $i$-th sample, $\epsilon_l$ and $\epsilon_h$ represent the low and high thresholds used to categorize accuracy tiers and prevent mathematical errors.  

\noindent \textbf{Accuracy Guidance Term.} When the accuracies of both modes reside within the same tier, their values exhibit close proximity. Given that accuracy is computed from a few discrete small samples, inherent randomness is present. Consequently, we regard the model's performance as comparable across modes under such conditions. Solely relying on the accuracy ratio to determine the weights $w_k$ may introduce excessive random disturbances during training, potentially causing oscillatory behavior in model optimization. Therefore, we design the accuracy guidance term $\mathbb{A}$ shown in~\cref{eq:adr_a_term}, which activates exclusively when the accuracies of both modes reside within the same tier, when \texttt{isSameTier}(·) is true. This term aims to: (1) encourage the model to engage in reasoning for accuracy improvement when both modes exhibit low-tier accuracy, and (2) avoid reasoning to reduce token consumption when both operate at medium or high tiers.

\begin{equation}
     \mathbb{A}(acc, k) = 
     \begin{cases} 
        \phi(k) \, * \, \psi(acc_k) , & \text{if \; \texttt{isSameTier}}(acc), \\ 
        0, & \text{otherwise.} 
    \end{cases} 
\label{eq:adr_a_term}
\end{equation}

\begin{align*}
\text{where:} & \\
& \phi(k) = \begin{cases} 
        1 , & \text{if} \; k=tk, \\ 
        -1, & \text{otherwise.} 
    \end{cases} \\
& \psi(acc) = \begin{cases} 
        1 , & \text{if} \; acc_k \ge \epsilon_h,  \\ 
        -1, & \text{otherwise.}
    \end{cases} \\
\end{align*}

\noindent \textbf{Propensity Guidance Term.} A critical challenge arises when mode selection becomes imbalanced during sampling. Consider a scenario where the model generates 16 samples for a given question: 15 samples select thinking mode with 13/15 accuracy (86.7\%), while only 1 sample selects non-thinking mode with 1/1 accuracy (100\%). The inclusion of the accuracy guidance term in our Decision Feedback steers model optimization toward non-thinking mode under such conditions. However, this conclusion is misleading. The model has already developed a strong preference for the thinking mode (15 of 16 samples), indicating that thinking is likely the right strategy for this query. Moreover, the 100\% accuracy of the non-thinking mode comes from a single sample, making it unreliable due to the small sample size. Forcing the optimization toward non-thinking mode based on this limited evidence would introduce large random fluctuations, potentially disrupting training through unstable gradients. To address this issue, we introduce a propensity guidance term $\mathbb{P}$ shown in~\cref{eq:adr_p_term} that accounts for the imbalance of the sample distribution and reduces noise from unreliable comparisons, optimizing the model toward it`s propensity when the model exhibits a propensity and both modes achieve high accuracy.

\begin{equation}
\begin{aligned}
& \mathbb{P}(acc,\rho,k,\Delta_{\rho}) = \\
& \begin{cases} 
        10 * \rho_k , & \text{if} \; |\rho_{tk} - \rho_{ntk}| >= \Delta_{\rho} \; \& \; \texttt{HighTier}(acc), \\ 
        0, & \text{otherwise.} 
    \end{cases} 
\end{aligned}
\label{eq:adr_p_term}
\end{equation}

\noindent where \texttt{HighTier}(·) identifies whether both thinking and non-thinking modes achieve high accuracy.

\noindent \textbf{Optimization.} For training stability and numerical safety, we normalize the mode-specific weights $w_k$ via~\cref{eq:adr_norm}, scaling both values to the closed interval [1, 2].

\begin{equation}
w_k^{norm} = clip \left ( \frac{w_k}{\textrm{min}(w_{tk}, w_{ntk}) + 10^{-6}}, 1, 2  \right )
\label{eq:adr_norm}
\end{equation}

In summary, the proposed Decision Feedback steers model optimization across multiple scenarios as detailed in~\cref{tab:adr_settings}. Through this carefully designed guidance framework, the model adaptively determines reasoning engagement per query while maintaining accuracy, eliminating reliance on manually enforced difficulty-based heuristics that rigidly prescribe reasoning behavior.

\begin{table}[t]
\centering
\caption{Adaptive decision reward condition settings. $\downarrow$ denotes the accuracy in low-tier, - indicates medium-tier, and $\uparrow$ represents high-tier.}
\begin{tabular}{c|c|c}
\hline
$Acc_{tk}$ & $Acc_{ntk}$ & Optimization Guidance \\
\hline
$\downarrow$ & $\downarrow$ & Thinking \\
$\downarrow$ & $-$ & Non-Thinking \\
$\downarrow$ & $\uparrow$ & Non-Thinking \\
$-$ & $\downarrow$ & Thinking \\
$-$ & $-$ & Non-Thinking \\
$-$ & $\uparrow$ & Non-Thinking \\
$\uparrow$ & $\downarrow$ & Thinking \\
$\uparrow$ & $-$ & Thinking \\
$\uparrow$ & $\uparrow$ &  
$\begin{cases} 
\text{Propensity}, & \text{if has propensity}, \\
\text{Non-Thinking}, & \text{otherwise.}
\end{cases}$\\
\hline
\end{tabular}
\label{tab:adr_settings}
\end{table}

\section{AdaThinking-Doc}

\noindent \textbf{Data Distribution.} To facilitate the training of AdaThinking-E, we present a diverse document understanding dataset named AdaThinking-Doc composed of both simple and complex scenarios. It encompasses two broad categories: 1) Information Extraction Scenarios. This includes datasets like DocVQA~\cite{mathew2021docvqa}, InfoVQA~\cite{singh2019towardstextvqa}, TextVQA~\cite{singh2019towardstextvqa}, STVQA~\cite{biten2019scene}, OCRVQA~\cite{mishra2019ocr}, DeepForm~\cite{svetlichnaya2020deepform}, FUNSD~\cite{jaume2019funsd}, SROIE~\cite{huang2019icdar2019sroie}, POIE~\cite{kuang2023visualpoie}, EST-VQA~\cite{wang2020general}, IAM~\cite{marti2002iam}, KLC~\cite{stanislawek2021kleister}, WTQ~\cite{pasupat2015compositionalwtq}, etc., where questions are directly posed about existing information in images, requiring no complex reasoning. 2) Reasoning Scenarios. This includes datasets such as ChartQA~\cite{masry2022chartqa}, CharXiv~\cite{wang2024charxiv}, DVQA~\cite{kafle2018dvqa}, IconQA~\cite{lu2021iconqa}, AI2D~\cite{kembhavi2016diagramworthdozenimagesai2d}, FigureQA~\cite{kahou2017figureqa}, PlotQA~\cite{methani2020plotqa}, A-OKVQA~\cite{schwenk2022okvqa}, ScienceQA~\cite{lu2022learnexplainmultimodalreasoningsqa}, which involve reasoning logic.

\noindent \textbf{Data Cleaning and Construction.} In Sec. 4.1 of the main text, we utilized Seed1.5-VL~\cite{guo2025seed15vltechnicalreport} to filter out 200,000 high-precision data. During this process, both the query and answer are input into Seed1.5-VL, which first determines whether the answer is correct. For correct answers, it generates a thought process in a specified format. Subsequently, we extract 80\% of the data from each source dataset~\cite{mathew2021docvqa,mathew2022infographicvqa,singh2019towardstextvqa,biten2019scene,mishra2019ocr,svetlichnaya2020deepform,jaume2019funsd,huang2019icdar2019sroie,kuang2023visualpoie,wang2020general,marti2002iam,stanislawek2021kleister,pasupat2015compositionalwtq,masry2022chartqa,wang2024charxiv,kafle2018dvqa,lu2021iconqa,kembhavi2016diagramworthdozenimagesai2d,kahou2017figureqa,methani2020plotqa,schwenk2022okvqa,lu2022learnexplainmultimodalreasoningsqa} to construct a cold start dataset, AdaThinking-Doc. In AdaThinking-Doc, each query generates both a thinking-type response and a Non-Thinking-type response. To ensure balanced patterns, entries and their corresponding thinking and Non-Thinking responses judged by GPT-4o-mini~\cite{openai2024gpt4ocard} as having an unreasonable reasoning process are discarded. Ultimately, 196,000 valid queries were retained. 

\noindent \textbf{Thinking Content.} Regarding the design of \texttt{thinking content}, we employ a step-by-step thinking mode, which includes a clear logical reasoning process and an effective problem-solving conclusion. For each step of thinking, it is necessary to first summarize the goal of the current step and then provide a detailed reasoning process. The specific format is as follows:
\begin{tcolorbox}[colframe=black, colback=white, boxrule=1pt, arc=4pt, width=1.\textwidth, left=0pt, right=0pt, top=0pt, bottom=0pt]

\noindent \textbf{\textsc{Thinking Content:}}

\noindent \texttt{\textbf{Step 1:}\{3-10 words summarise the goal for the current step\}}

\noindent \texttt{\{detailed reasoning process\}}

\noindent \texttt{\textbf{Step 2:}\{3-10 words summarise the goal for the current step\}}

\noindent \texttt{\{detailed reasoning process\}}

\noindent \texttt{\textbf{...}}

\noindent \texttt{\textbf{Step N:}\{3-10 words summarise the goal for the current step\}}

\noindent \texttt{\{detailed reasoning process\}}

\noindent \texttt{\textbf{Summary:}\{the conclusion to the query\}}
\end{tcolorbox}

\section{More Results}

\begin{table}[htbp]
    \centering
    \caption{Comparison of AdaThinking-E's performance during the cold start and RL stages across different benchmarks.}
    \label{tab:supp_stage}
    \resizebox{1.\textwidth}{!}{
    \begin{tabular}{c|cc|ccc|ccc|cc}
    \toprule
    Model & SFT & RL & \textbf{DocVQA} & \textbf{InfoVQA} & \textbf{Text\textsubscript{Val}} & \textbf{ChartQA} & \textbf{CharXiv$_{RQ}$} & \textbf{CharXiv$_{DQ}$} & \textbf{OCR-Bench} & \textbf{OCR-Reasoning} \\
    \midrule
    QwenVL2.5-8B~\cite{bai2025qwen25vltechnicalreport} & - & - & 95.7 & 82.6 & 84.9 & 87.3 & 42.5 & 73.9 & 86.4 & 15.7 \\
    QwenVL3-8B~\cite{bai2025qwen3vltechnicalreport} & - & - & 95.3 & 86.0 & 78.7 & 88.6 & 53.0 & 85.9 & 81.9 & 48.4  \\
    \midrule
    \multirow{2}{*}{\shortstack{AdaThinking-E \\ (Qwen2.5)}} & $\checkmark$ &  & 95.9 & 82.1 & \underline{84.5} & 87.6 & 47.6 & 78.2 & 86.7 & 21.6 \\ 
    & $\checkmark$ & $\checkmark$ & \underline{96.3} & 82.9 & \textbf{85.3} & \underline{89.1} & \underline{56.7} & 83.5 & \underline{88.7} & 26.5  \\
    \midrule
    \multirow{2}{*}{\shortstack{AdaThinking-E \\ (Qwen3)}} & $\checkmark$ &  & 95.8 & \underline{85.9} & 79.9 &  \underline{89.1} & 54.3 & \underline{86.1} & 84.9 & \underline{48.8} \\ 
    & $\checkmark$ & $\checkmark$ & \textbf{96.4} & \textbf{86.3} & 82.8 & \textbf{90.1} & \textbf{57.3} & \textbf{86.6} & \textbf{89.8} & \textbf{49.7} \\
    \bottomrule
    \end{tabular}
    }
\end{table}

\noindent \textbf{Training phase ablations.} In~\cref{tab:supp_stage}, we supplemented the performance of AdaThinking-E at different stages of training on different benchmarks~\cite{mathew2021docvqa,mathew2022infographicvqa,singh2019towardstextvqa,masry2022chartqa,wang2024charxiv,liu2023hiddenocrbench,huang2025ocr}. Compared to QwenVL2.5-7B~\cite{bai2025qwen25vltechnicalreport}, AdaThinking-E-Qwen2.5 in the cold start stage showed improvements of 0.6\%, 5.1\%, and 5.9\% on reasoning-oriented benchmarks such as ChartQA~\cite{masry2022chartqa}, CharXiv$_{RQ}$~\cite{wang2024charxiv}, and OCR-Reasoning~\cite{huang2025ocr}, respectively, demonstrating the effectiveness of the AdaThinking-Doc construction. However, in simpler scenarios like InfoVQA~\cite{mathew2022infographicvqa} and TextVQA~\cite{singh2019towardstextvqa}, excessive reasoning led to a decrease in performance by 0.5\% and 0.4\%, respectively. Through the RL stage, AdaThinking-E-Qwen2.5 fully developed adaptive thinking capabilities. Compared to the cold start stage, AdaThinking-E-Qwen2.5 exhibited outstanding performance in both simple and reasoning scenarios, with an average improvement of 3.1\%. An intriguing observation is that AdaThinking-E-Qwen3 shows marginal gains on sophisticated benchmarks like CharXiv and OCR-Reasoning relative to AdaThinking-E-Qwen2.5, yet outperforms it on simple data. This suggests that our approach provides non-thinking models with better solutions for complex problems while mitigating over-thinking-induced hallucinations in thinking models when handling simple data. Such results demonstrate the essential role of adaptive thinking for document understanding and confirm the effectiveness of AdaThinking-E.

\begin{figure}[t]
    \centering
    \includegraphics[width=1.\linewidth]{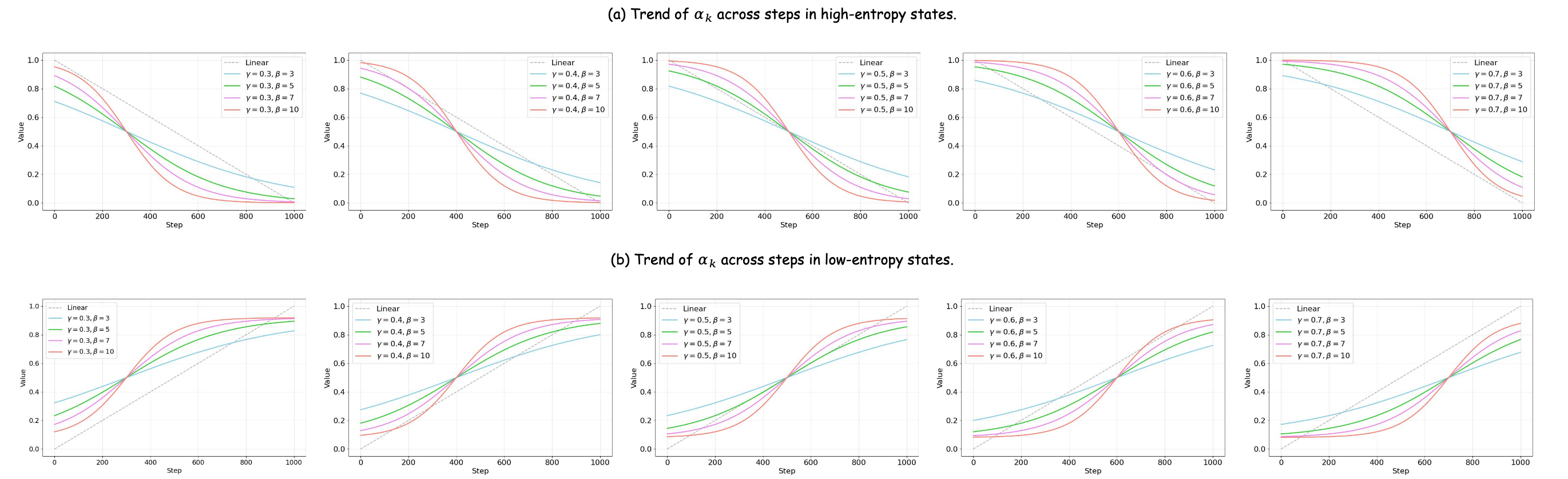}
    \caption{Evolution of $\alpha_k$ relative to step size under different conditions.}
    \label{fig:alpha-k}
\end{figure}

\noindent \textbf{Analysis of Dynamic Scoring Mechanism.} ~\cref{fig:alpha-k} illustrates the trends of the dynamic scoring mechanism across training steps under various configurations of $\gamma$ and $\beta$. As $\beta$ increases, the distinction between the exploration and convergence phases becomes more pronounced, enabling the model to better perceive state transitions. Furthermore, $\gamma$ serves as the critical inflection point governing the transition from exploration to convergence. Beyond this point, the weight assigned to low-entropy states gradually surpasses that of high-entropy states. A premature transition (small $\gamma$) prevents the model from adequately assessing whether a query requires thinking, leading to "lazy", homogenized outputs. Conversely, a delayed transition causes the model to lack "confidence" in its decisions. Ultimately, we achieve optimal performance across comprehensive benchmarks with $\gamma=0.4$ and $\beta=10$.

\section{Case Study}

\begin{figure}[htbp]
    \centering
    \includegraphics[width=0.6\linewidth]{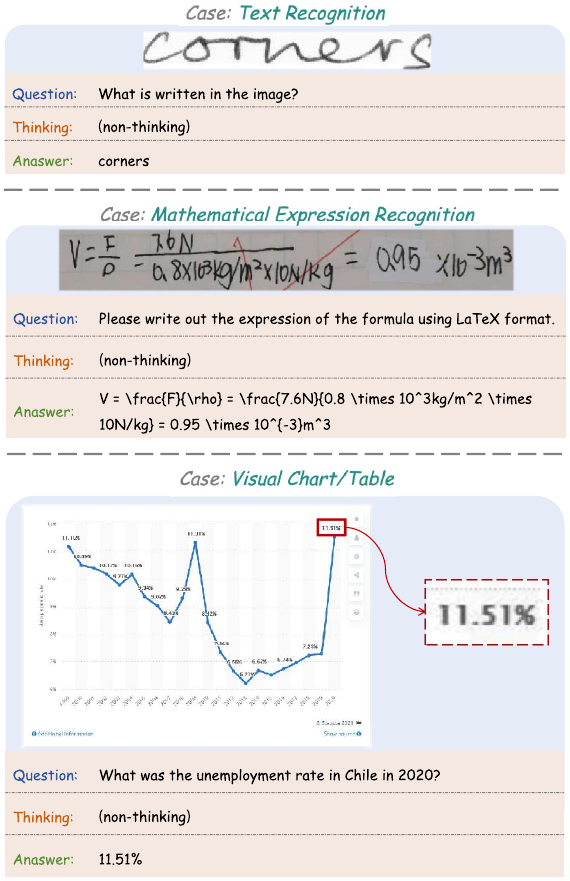}
    \caption{Examples of AdaThinking-E in non-thinking mode.}
    \label{fig:supp-ntk-1}
\end{figure}

\begin{figure}[t]
    \centering
    \includegraphics[width=0.8\linewidth]{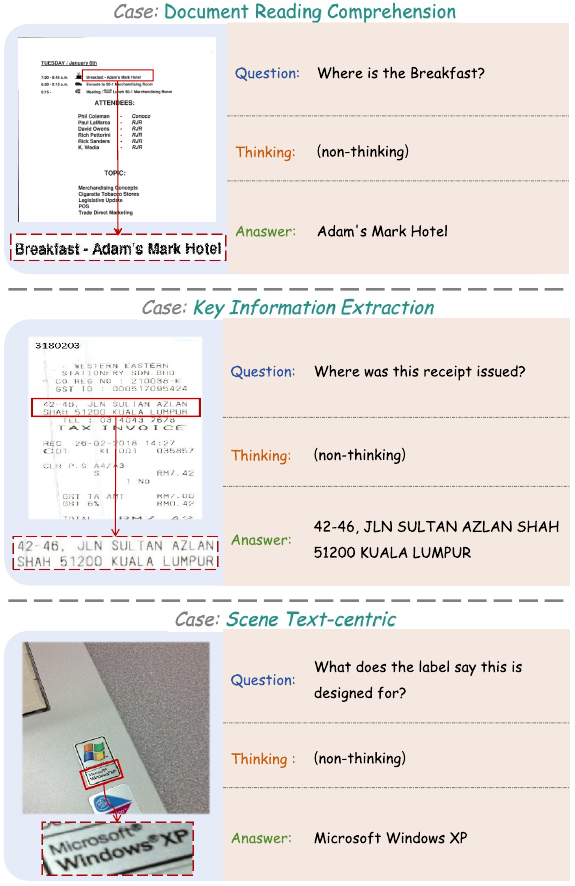}
    \caption{Examples of AdaThinking-E in non-thinking mode.}
    \label{fig:supp-ntk-2}
\end{figure}

\begin{figure}[t]
    \centering
    \includegraphics[width=0.9\linewidth]{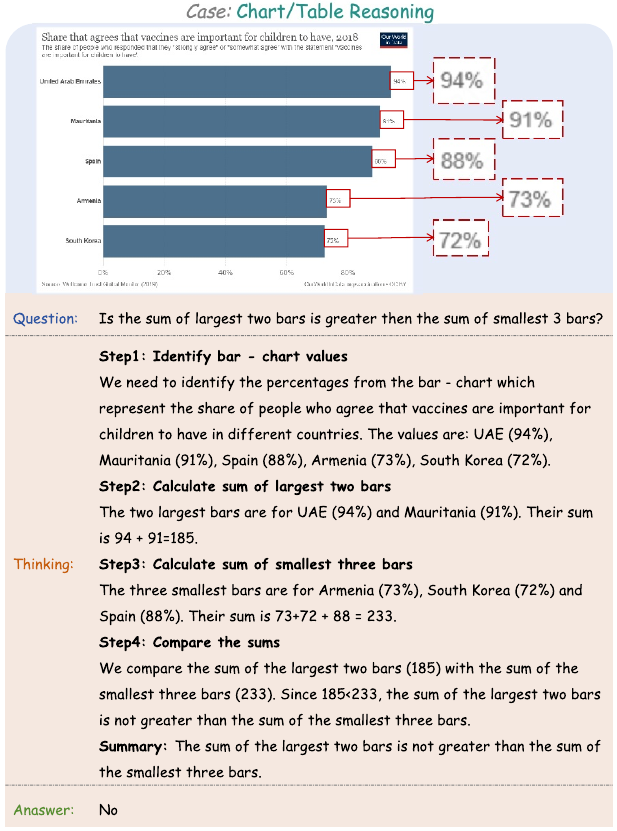}
    \caption{An example of AdaThinking-E in thinking mode.}
    \label{fig:supp-tk-1}
\end{figure}

\begin{figure}[t]
    \centering
    \includegraphics[width=0.9\linewidth]{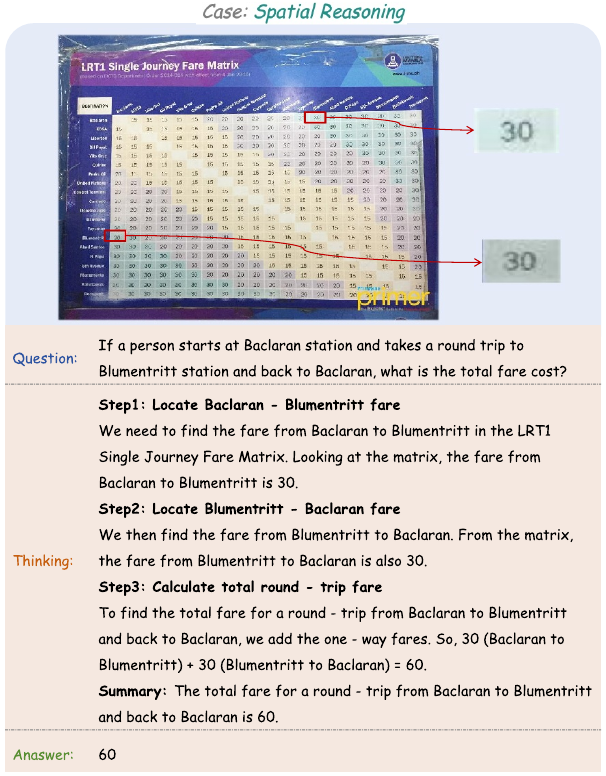}
    \caption{An example of AdaThinking-E in thinking mode.}
    \label{fig:supp-tk-2}
\end{figure}

\begin{figure}[t]
    \centering
    \includegraphics[width=.9\linewidth]{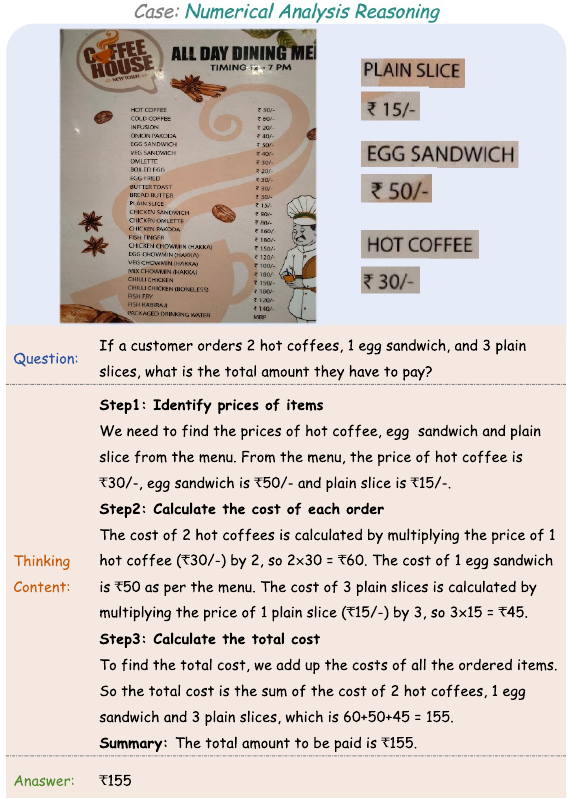}
    \caption{An example of AdaThinking-E in thinking mode.}
    \label{fig:supp-tk-3}
\end{figure}

\begin{figure}[t]
    \centering
    \includegraphics[width=.9\linewidth]{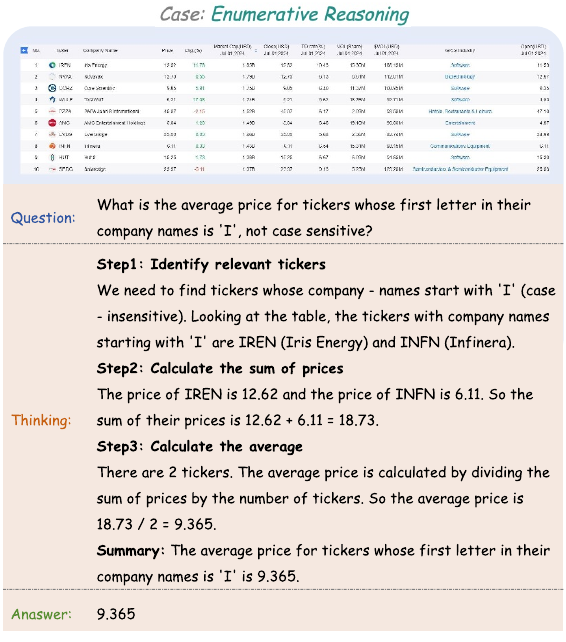}
    \caption{An example of AdaThinking-E in thinking mode.}
    \label{fig:supp-tk-4}
\end{figure}

\begin{figure}[t]
    \centering
    \includegraphics[width=0.85\linewidth]{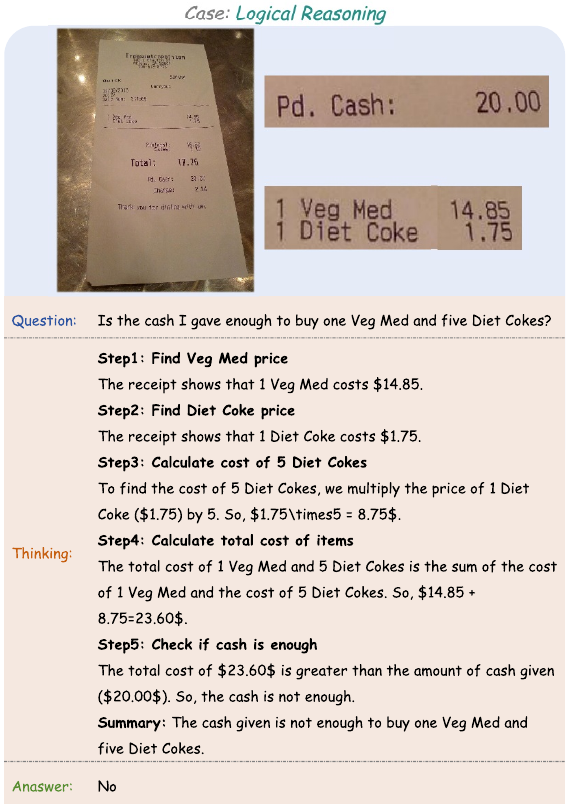}
    \caption{An example of AdaThinking-E in thinking mode.}
    \label{fig:supp-tk-5}
\end{figure}

\begin{figure}[t]
    \centering
    \includegraphics[width=0.7\linewidth]{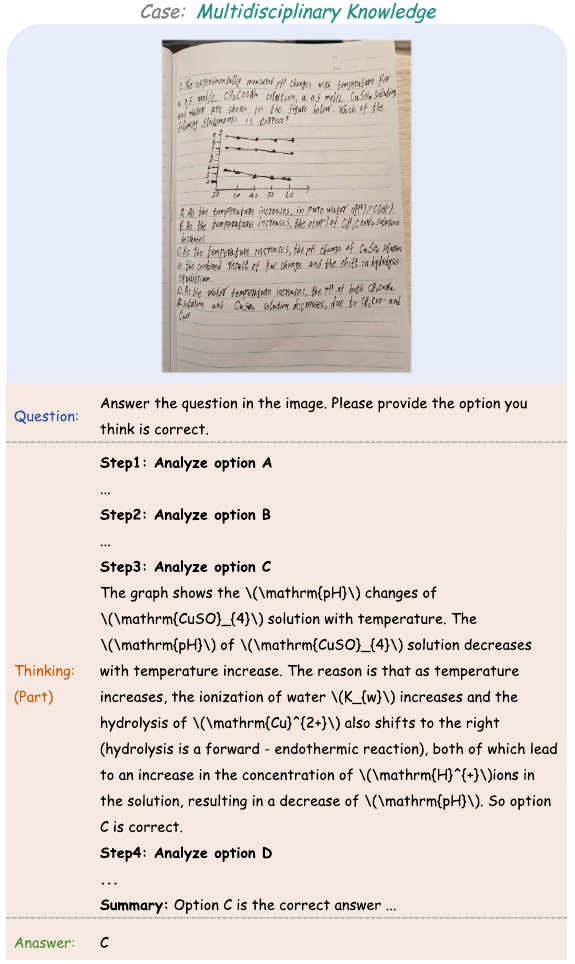}
    \caption{An example of AdaThinking-E in thinking mode.}
    \label{fig:supp-tk-6}
\end{figure}

In~\cref{fig:supp-ntk-1,fig:supp-ntk-2,fig:supp-tk-1,fig:supp-tk-2,fig:supp-tk-3,fig:supp-tk-4,fig:supp-tk-5,fig:supp-tk-6}, we demonstrate the switching between thinking and non-thinking modes of AdaThinking-E across different task types. As illustrated in~\cref{fig:supp-ntk-1,fig:supp-ntk-2}, AdaThinking-E opts for the non-thinking mode in straightforward scenarios such as OCR, information extraction, and visual question answering, providing quick responses. Conversely, in the logical reasoning and analysis scenarios depicted in~\cref{fig:supp-tk-1,fig:supp-tk-2,fig:supp-tk-3,fig:supp-tk-4,fig:supp-tk-5,fig:supp-tk-6}, AdaThinking-E chooses the thinking mode, engaging in step-by-step reasoning before arriving at the final conclusion. By switching between different modes, AdaThinking-E is able to handle various simple or complex document scenarios, ensuring efficient responses while maintaining accuracy.

\clearpage

\end{document}